\documentclass[10pt,twocolumn,letterpaper]{article}
\usepackage{placeins}
\usepackage{cvpr24/cvpr}              

\usepackage[dvipsnames]{xcolor}

\usepackage{booktabs}
\usepackage{tabularx} 
\usepackage{rotating}
\usepackage{makecell} 

\newcommand{\loss}[1]{\mathcal{L}_\text{#1}}

\definecolor{cvprblue}{rgb}{0.21,0.49,0.74}
\usepackage[pagebackref,breaklinks,colorlinks,citecolor=cvprblue]{hyperref}

\def\paperID{19} 
\def\confName{CVPR}
\def\confYear{2024}

\title{PC-LoRA: Low-Rank Adaptation for Progressive Model Compression with Knowledge Distillation
\vspace{-0.4cm}}

\author{
Injoon Hwang\textsuperscript{1}\thanks{These authors contributed equally to this work.}, \
HaeWon Park\textsuperscript{1}\footnotemark[1], \
Youngwan Lee\textsuperscript{2,3}\footnotemark[1], \
Jooyoung Yang\textsuperscript{1}, \
SunJae Maeng\textsuperscript{1} \\
\vspace{-0.8cm}
\small
\textsuperscript{1}C-LoRA LAB, MODULABS \
\textsuperscript{2}ETRI \
\textsuperscript{3}KAIST \\
}

\begin{document}
\maketitle

\begin{abstract}
Low-rank adaption (LoRA) is a prominent method that adds a small number of learnable parameters to the frozen pre-trained weights for parameter-efficient fine-tuning. Prompted by the question, ``Can we make its representation enough with LoRA weights solely at the final phase of finetuning without the pre-trained weights?'' In this work, we introduce Progressive Compression LoRA~(PC-LoRA), which utilizes low-rank adaptation (LoRA) to simultaneously perform model compression and fine-tuning. The PC-LoRA method gradually removes the pre-trained weights during the training process, eventually leaving only the low-rank adapters in the end. Thus, these low-rank adapters replace the whole pre-trained weights, achieving the goals of compression and fine-tuning at the same time. Empirical analysis across various models demonstrates that PC-LoRA achieves parameter and FLOPs compression rates of 94.36\%/89.1\% for vision models, e.g., ViT-B, and 93.42\%/84.2\% parameters and FLOPs compressions for language models,  e.g., BERT. 
\end{abstract}


\section{Introduction}

Ever since pre-trained Transformer~\cite{vaswani2023attention} models were introduced, they have shown outstanding effectiveness in a range of tasks within Natural Language Processing (NLP)~\cite{brown2020language, devlin2019bert} and Computer Vision (CV)~\citep{dosovitskiy2021image, carion2020endtoend, wang2021pyramid} tasks. 
However, their substantial size and the high computational demands pose difficulties in both deployment and fine-tuning.

To address these challenges, several parameter-efficient fine-tuning methods have been introduced, including Prefix Tuning~\cite{li2021prefixtuning}, Prompt Tuning~\citep{Lester2021ThePO}, P-Tuning~\cite{Liu2021GPTUT}, adapters~\cite{houlsby2019parameter}, and Low-Rank Adaptation (LoRA)~\cite{hu2021lora}.
Specifically, LoRA employs \textit{trainable} low-rank matrices within transformer layers, which drastically cuts down the number of trainable parameters for fine-tuning. 
However, LoRA is only memory efficient during fine-tuning; when a model fine-tuned with LoRA is used for inference, it offers no advantages over the original pre-trained model.

\begin{figure}[t]
\centering
\includegraphics[width=\linewidth]{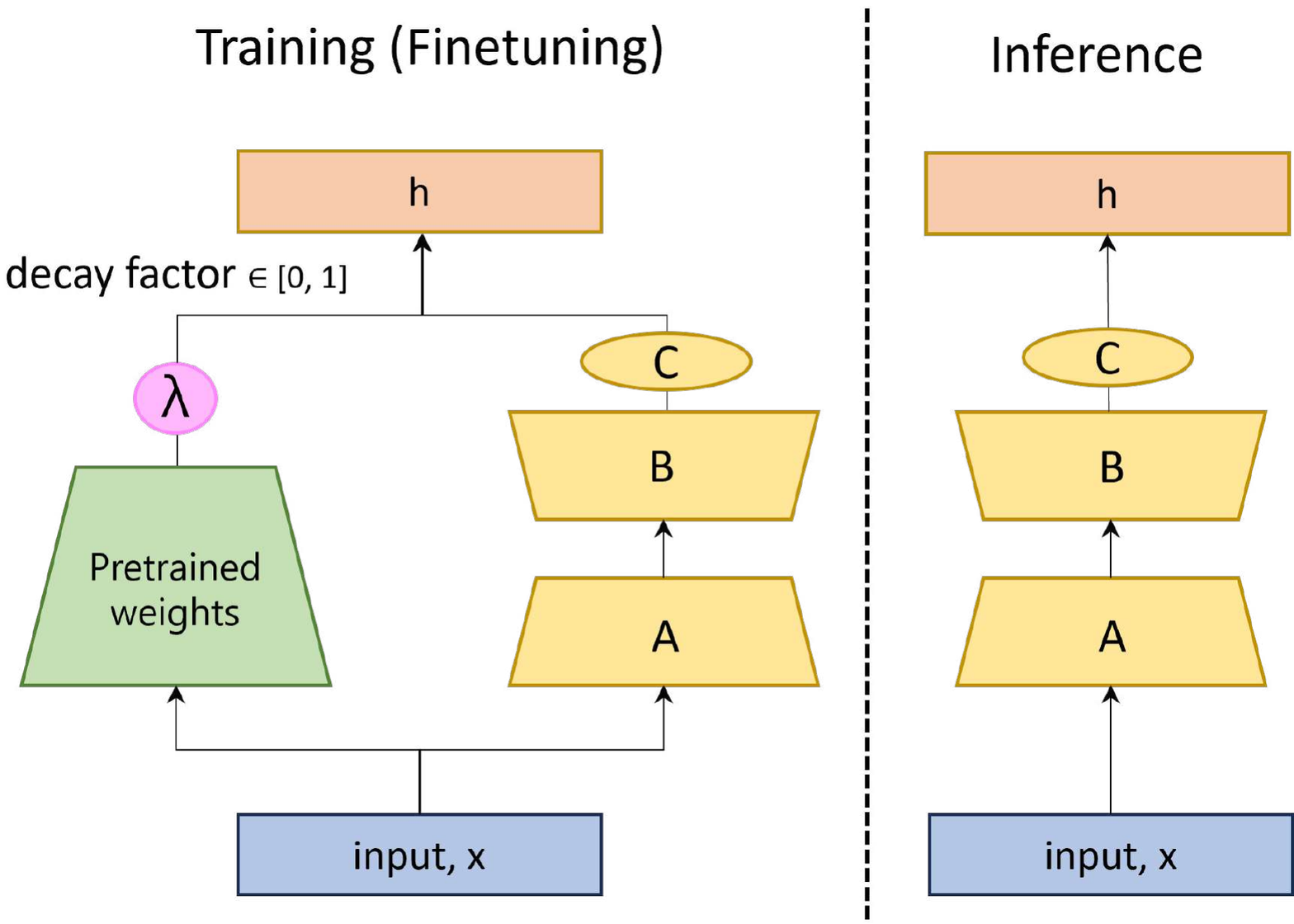}
\caption{The overall diagram of the PC-LoRA method. At each training step, the pre-trained weights and bias gradually decay according to a decay factor $\lambda$, and eventually disappear and only the Low-Rank Adapter corresponding weights $A$, $B$ and bias $C$ remain.}
\vspace{-0.3cm}
\label{fig:PC-LoRA}
\end{figure}

Based upon the foundation of LoRA, we question ``\textit{Can we achieve sufficient representation using only LoRA weights in the final phase of fine-tuning, without relying on pre-trained weights?}''
To find the answer to this question, we propose a \textbf{Progressive Compression-LoRA~(PC-LoRA)} method, which gradually reduces the pre-trained weights~ (i.e., the base weights) during fine-tuning until they are completely removed as shown in Figure \ref{fig:PC-LoRA}.

In order to learn the representation of the pre-trained model within the low-rank adapter, the PC-LoRA method progressively attenuates the output of the pre-trained weights during the fine-tuning phase.
This decay will result in the loss of information, which prompts the low-rank adapter to increasingly attempt to compensate. Through this process, the representations of the pre-trained model will effectively flow into the low-rank adapter.
Therefore, by the end of the training, the LoRA weights may contain both the representations of the original weights and the updates made for fine-tuning.
This capability demonstrates how PC-LoRA methodically achieves the dual objectives of low-rank compression and parameter-efficient fine-tuning.

\begin{figure}[t]
\hfill
\centering
\includegraphics[width=\linewidth]{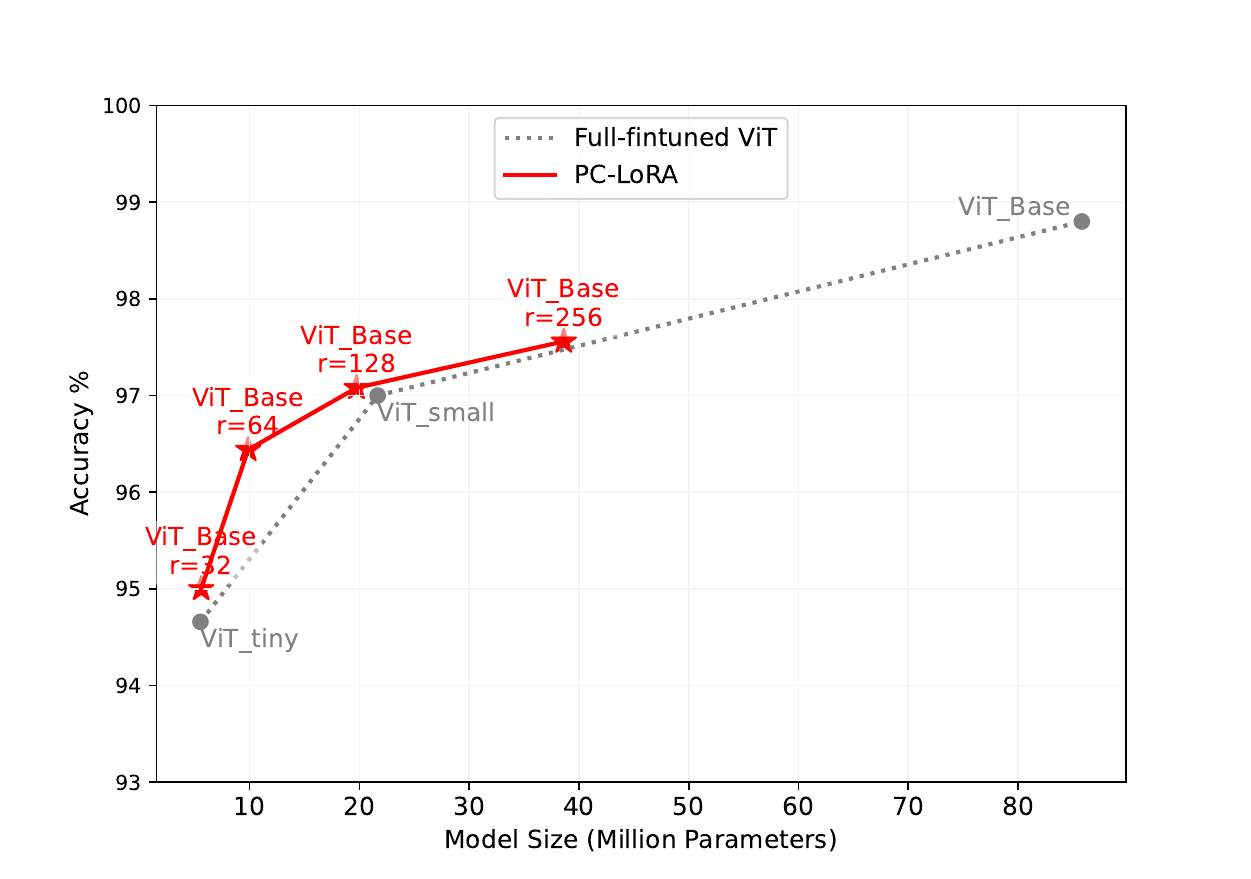}
\vspace{-0.6cm}
\caption{The performance comparisons based on different compression ratios of PC-LoRA using ViT-B~\cite{dosovitskiy2021image} compared to the fully finetuned ViT-B on CIFAR-10. 
}
\vspace{-0.3cm}
\label{fig:vit_line_plot}
\end{figure}

Our PC-LoRA drastically reduces model parameters and computes significantly with a slight accuracy drop, achieving up to 94.1\%p parameter reduction and 89.1\%p FLOPs decrease for vision models \eg, ViT-B~\cite{dosovitskiy2021image}, and up to 93.5\%p parameter and 84.2\%p FLOPs reduction for language models, \eg, BERT~\cite{devlin2019bert}.
\cref{fig:vit_line_plot} shows several compressed models by simply adjusting low rank~($r$) in the LoRA weights, demonstrating that our PC-LoRA exhibits flexibility and scalability.
In addition to reducing the model budget, PC-LoRA provides further advantages. Its layer-wise approach allows the application to any model with linear layers, enhancing scalability. Moreover, PC-LoRA is compatible with other methods like quantization~\cite{jacob2017quantization, dettmers2023qlora}, enabling combined use to improve model efficiency.

\section{PC-LoRA Method}

Our method, termed Progressive Compression with Low-Rank Adaptation (PC-LoRA), is designed to incrementally compress a model by diminishing and eventually removing the influence of pre-trained weights throughout the training process. 
In this approach, both pre-trained model weights and low-rank adapter weights are initially used for the output computation. As the training progresses, the pre-trained model weights gradually fade away according to a decay factor and are eventually removed at the end of the training, leaving only the low-rank adapters. 
The overall concept is illustrated in Figure \ref{fig:PC-LoRA}.

\noindent
\textbf{PC-LoRA Layer}:
Similar to the LoRA~\cite{hu2021lora} method, our PC-LoRA method also attaches low-rank adapters to linear layers.
In PC-LoRA, the layers consist of the weight and bias from the pre-trained model, complemented by two low-rank adapter weights, \(A\) and \(B\) with a rank~($r$), which replace the pre-trained model’s weight \(W\). Additionally, weight \(C\) is used to substitute for the bias.

\FloatBarrier

\noindent
\subsection{Training Configuration}
The model's output at each layer is calculated by adding the decayed pre-trained model output (\(D_i\)) with the low-rank adapter output (\(L_i\)), where \(i\) denotes the index of the specific LoRA layer. 
A decay factor \(\lambda\) is used to gradually lessen the impact of the pre-trained model output, decreasing from 1 to 0. 
The details of the decay factor scheduling will be described in~\cref{sec:decay}.
While training, the final output (\(F_i\)) is computed as:

\vspace{-0.2cm}

\begin{align}
    D_i &= \lambda W_i(x_i), \tag{1} \label{eq:1} \\
    L_i &= B_i(A_i(x_i)) + C_i, \tag{2} \label{eq:2} \\
    F_i &= D_i + L_i. \tag{3} \label{eq:3}
\end{align}

During training, \(A_i\), \(B_i\) and \(C_i\) are trainable, while the pre-trained model weights (\(W_i\)) and biases remain unchanged. The bias is not included in Equation \ref{eq:1} for simplicity. 
Initially, \(A_i\) is initialized with a random Gaussian distribution, and both \(B_i\) and \(C_i\) start at zero, meaning \(B_iA_i\) starts at zero as well.

After the completion of training, the output from the pre-trained model is completely eliminated by the decay factor, resulting in the forward pass being represented by the following equation: \( F_i = L_i \), indicating that only the low-rank adapter is used. 
Note that when we trained exclusively with the low-rank adapter, we observed no improvement in the model's performance; specifically, it achieved only 10\% accuracy on CIFAR-10 and 50\% accuracy on IMDb, which demonstrates that our method is more effective than the simple layer-pruning method.


The PC-LoRA method is optimized according to the loss term as follows:
\begin{align}
    \loss{total} &= \alpha \cdot  \loss{task}(y_i, \hat{y}_i) + (1 - \alpha) \cdot  \loss{featKD}(F_S, F_T) \tag{4} \label{eq:4}
\end{align}
The total loss \(\loss{total}\) is defined by combining the task loss, \(\loss{task}\), with the feature-based knoweldge distillation loss, \(\loss{featKD}\).
\(\loss{task}\) represents the loss for fine-tuning on a downstream task, such as cross-entropy loss for classification tasks with labels. 
\(\loss{featKD}\) is computed as the Mean Squared Error (MSE) between the intermediate features of the student model \(F_S\) and those of the teacher model \(F_T\) as below:
\begin{align}
     \loss{featKD}(F_S, F_T) &= \frac{1}{M} \sum_{m=1}^{M} \text{MSE}(F_{S_m}, F_{T_m}),  \tag{5} \label{eq:5} 
\end{align}
where \(M\) is the number of PC-LoRA applied layers. The student \(S\) is the model with PC-LoRA applied, which includes the decayed pre-trained model and low-rank adapters, while the teacher \(T\) is the original pre-trained model. 
For the layers where PC-LoRA is applied, incorporating the difference between the intermediate feature maps of \(S\) and \(T\) into the loss term acts as a \textit{regularization}. Since the teacher model \(T\) remains in its original, un-finetuned state, adding this term helps to prevent the \(S\), which will eventually retain only the low-rank adapters, from overfitting to the downstream task while training. The ablation study for the effect of $\loss{featKD}$ is conducted in~\cref{tab:fd_loss_performance}. 
Please see~\cref{appendix:loss_scale_factor} for an ablation study of the loss scale factor, $\alpha$.

\vspace{-0.1cm}

\subsection{Decay Factor Scheduler}\label{sec:decay}
\begin{figure}[t]
\centering
\includegraphics[width=0.7\linewidth]{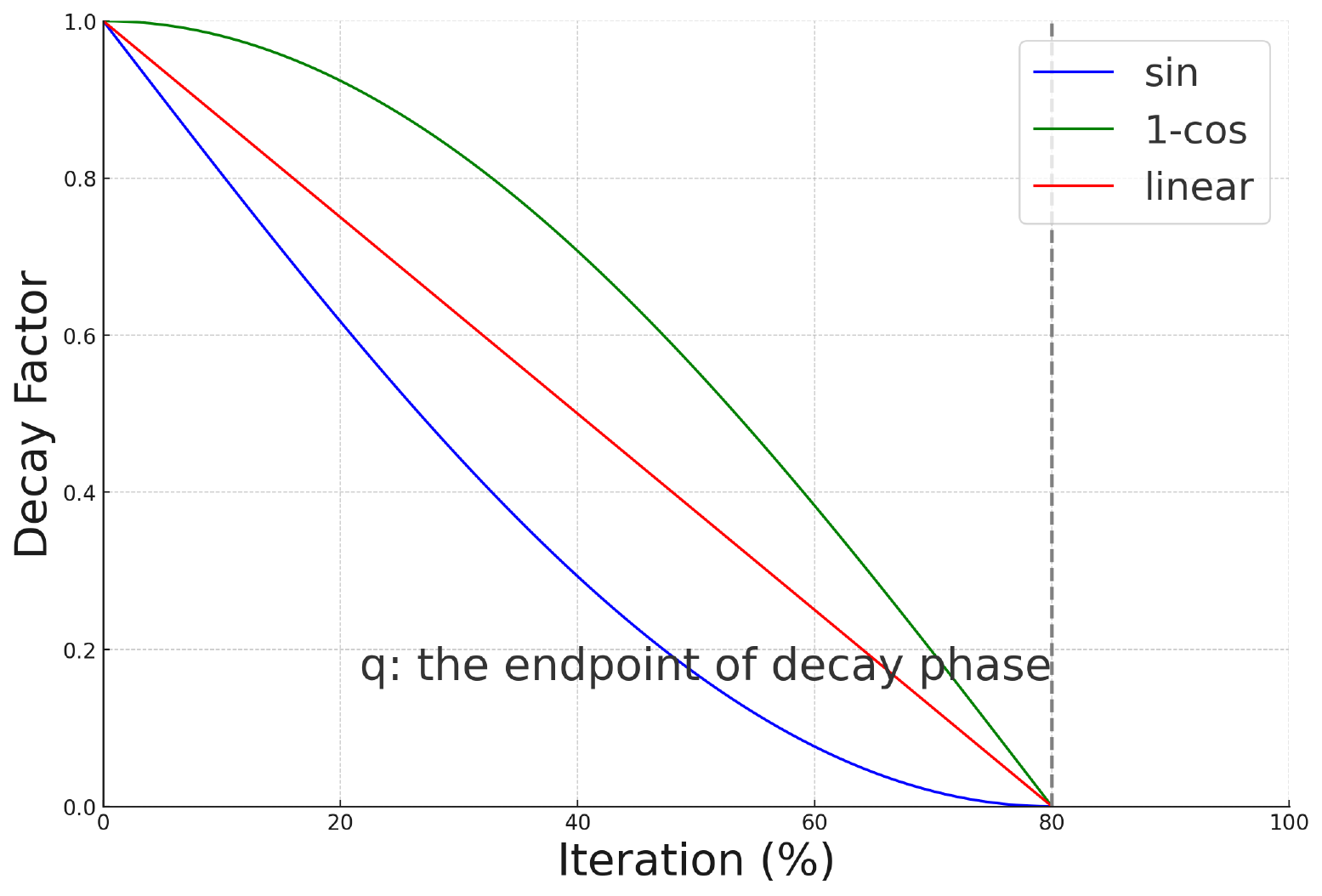}
\vspace{-0.35cm}
\caption{The three types of Decay Factor Scheduler: Sine, 1-Cosine, and Linear. As iterations progress, the decay factor decreases from 1 to 0, affecting the rate at which the original weight becomes less influential. Initially, a factor of 1 means the pre-trained model's weights are entirely preserved, while a factor of 0 indicates the complete transition to the new weights. 
}
\vspace{-0.3cm}
\label{fig:wds}
\end{figure}

This essential component of PC-LoRA manages the decay of the original model's weights to optimize compression and adaptation. It uses a decay factor \(\lambda \), which transitions from 1 to 0 based on a selected decay function. This process is detailed in the decay equation and graphically depicted in~\cref{fig:wds}.
The equation is as follows:

\vspace{-0.3cm}
\begin{align*}
\lambda(n) = \begin{cases}
1 - \text{Selected Decay function}(n, q) & \text{if } n < q, \\
0 & \text{if } n \geq q.
\end{cases}
\end{align*}
\vspace{-0.3cm}

\noindent
Here, \( q \) represents the endpoint of the decay phase, set as a proportion of the total iterations \( N \). The selected decay function dictates how \(\lambda(n) \) is reduced from 1, influencing the rate at which the influence of the original weights diminishes. Initially, \(\lambda(n) = 1\) indicates that the pre-trained model is fully intact, whereas \(\lambda(n) = 0\) signifies that the original weights have been completely phased out. The value \( q \) is ideally set between 40\% and 80\% of total iterations, within which range significant performance differences have not been observed.
From an ablation study of the decay function in~\cref{tab:decay_scheduler_performance}, we set \texttt{sine} as a default for all experiments.

\section{Experiments}

\subsection{Implementation Details}
In this paper, we investigated model performances in two benchmark tasks: image classification with the CIFAR-10~\cite{Krizhevsky2009LearningML} dataset and text classification using the IMDb~\cite{maas-etal-2011-learning} dataset. Training details are provided in Appendix \ref{appendix:a} and \ref{appendix:b}.  

We compared PC-LoRA with two other methods for fine-tuning pre-trained models. The first method, \texttt{Full Fine-Tuning (Full-FT)}, updates all parameters of the model. The second, \texttt{LoRA Fine-Tuning (LoRA-FT)}, incorporates LoRA but does not alter the embedding layers and uses a fixed rank of 32. The PC-LoRA approach uses the same configuration as the LoRA-FT method, focusing training only on layers modified for low-rank adjustments.

Furthermore, we evaluated how different ranks influence the compressed model size and performance, comparing with various sizes of ViT models, as illustrated in Figure \ref{fig:vit_line_plot}. Similarly, we extended our analysis to BERT models, as detailed in Appendix \ref{appendix:E}.

In our research, we conducted ablation studies on the PC-LoRA method, detailed in Appendix \ref{appendix:D}. These studies included exploring different types of Decay Factor Schedulers to determine their impact on performance. Additionally, we investigated the optimal ratio for feature-based knowledge distillation loss, denoted as the $\alpha$ value in Equation \ref{eq:4}, to enhance the accuracy of the compressed model.

\subsection{Main Results}

\begin{table*}[t]
    \centering
    \begin{minipage}[t]{0.45\textwidth}
        \centering
        \scalebox{0.80}{
        \begin{tabular}{llrrl}
        \toprule
        Model & Setting & Acc(\%) & GFLOPs & Param(M) \\
        \midrule
        ViT-B~\cite{dosovitskiy2021image}            & Full-FT    & 99.00 & 16.87 & 85.81 \\
                         & LoRA-FT    & 98.80 & 16.67 & 85.81 \\
                         & PC-LoRA    & 95.04 &  1.07 & 5.94 (94.36\%↓) \\
        \cmidrule{1-5}
        ViT-B/Clip~\cite{radford2021learning}       & Full-FT    & 98.07 & 16.87 & 85.81 \\
                         & LoRA-FT    & 98.14 & 16.67 & 85.81 \\
                         & PC-LoRA    & 94.12 &  1.07 & 5.94 (94.36\%↓) \\
        \cmidrule{1-5}
        ViT-B/DINO~\cite{caron2021emerging}       & Full-FT    & 97.97 & 16.87 & 85.81 \\
                         & LoRA-FT    & 98.86 & 16.67 & 85.81 \\
                         & PC-LoRA    & 95.20 &  1.07 & 5.94 (94.36\%↓) \\
        \bottomrule
        \end{tabular}
        }
        \captionof{table}{Comparison of PC-LoRA (r=32), Full-FT, and LoRA-FT methods by different pre-trained models on CIFAR10, measuring GFLOPs and parameters during inference. Parentheses indicate the percentage reduction in parameters with LoRA-FT compared to PC-LoRA.}
        \label{tab:cifar10}
    \end{minipage}
    \hfill
    \begin{minipage}[t]{0.45\textwidth}
        \centering
        \scalebox{0.80}{
        \begin{tabular}{llrrl}
        \toprule
        Model & Setting & Acc(\%) & GFLOPs & Param(M) \\
        \midrule
        BERT~\cite{devlin2019bert}     & Full-FT & 94.00 & 48.37 & 109.48 \\
                 & LoRA-FT & 94.01 & 48.37 & 109.48 \\
                 & PC-LoRA & 92.78 &  7.60 & 29.34 (93.42\%↓) \\
        \cmidrule{1-5}
        ELECTRA~\cite{clark2020electra}  & Full-FT & 95.33 & 48.37 & 109.48 \\
                 & LoRA-FT & 94.63 & 48.37 & 109.48 \\
                 & PC-LoRA & 92.86 &  7.60 & 29.34 (93.42\%↓) \\
        \cmidrule{1-5}
        RoBERTa~\cite{liu2019roberta}  & Full-FT & 94.67 & 48.37 & 124.65 \\
                 & LoRA-FT & 94.40 & 48.37 & 124.65 \\
                 & PC-LoRA & 92.59 &  7.60 & 44.50 (93.42\%↓) \\
        \bottomrule
        \end{tabular}
        }
        \captionof{table}{Comparison of PC-LoRA (r=32), Full-FT, and LoRA-FT methods on different pre-trained models with IMDb, measuring GFLOPs and parameters during inference. Parentheses indicate the percentage reduction in parameters with LoRA-FT compared to PC-LoRA, excluding embedding layers.}
        \label{tab:imdb}
    \end{minipage}
    \vspace{-0.2cm}
\end{table*}
\cref{tab:cifar10} \& \cref{tab:imdb} presents a comparative analysis of the performance of various vision and language models on the CIFAR10 and IMDb benchmarks, utilizing Full-FT, LoRA-FT, and the proposed PC-LoRA method. 
 The performance of Full-FT and LoRA-FT is similar. When compared to the LoRA-FT method, which uses the same number of parameters for training, the PC-LoRA method shows an average performance degradation of about -3.56\%p. 
 Despite such performance degradation, the final outcome of PC-LoRA is a compressed model, resulting in a reduction of 94.36\%p excluding the embedding layers, and an average 89.1\%p decrease in total GFLOPs in vision models. For NLP models, also excluding the embedding layer, the reductions are about 93.42\%p, and 84.2\%p in total GFLOPs. As the results indicate, the PC-LoRA method demonstrates a favorable trade-off on both CIFAR10 and IMDb benchmarks, by significantly reducing the GFLOPs and model parameters while only modestly compromising accuracy.

In Figure \ref{fig:vit_line_plot}, the performance of various ViT models, including compressed ViT Base models with PC-LoRA method at different ranks, along with ViT Base, Tiny, and Small models, is displayed.
The x-axis represents the model size, and the y-axis shows the test accuracy on CIFAR10. 
The points marked with stars indicate the performance of the PC-LoRA compressed models, which are comparable to Vit Small and Tiny.
Two key observations can be made from the results in Figure \ref{fig:vit_line_plot}. First, the models compressed using the PC-LoRA method outperform the ViT tiny and ViT small models despite having a similar model size. Second, the PC-LoRA method allows for compression by adjusting a factor rank, enabling the preservation of the original model structure while adjusting its size to the desired level. Therefore our method is capable of generating models not only at the size levels of ViT Tiny, Small, and Base but also at any desired model size.

Similarly, Figure \ref{fig:bert_graph} displays the performance of various BERT models, including compressed Bert Base models with PC-LoRA method at different ranks, along with BERT Base, Medium, and Small. Figure \ref{fig:bert_graph} shows that similar to the results in Figure \ref{fig:vit_line_plot}, PC-LoRA compressed models outperform similar-sized models. This consistency across different models demonstrates the robustness and versatility of applying the our method for both CV and NLP tasks.

\begin{figure}[h]
\centering
\includegraphics[width=\linewidth]{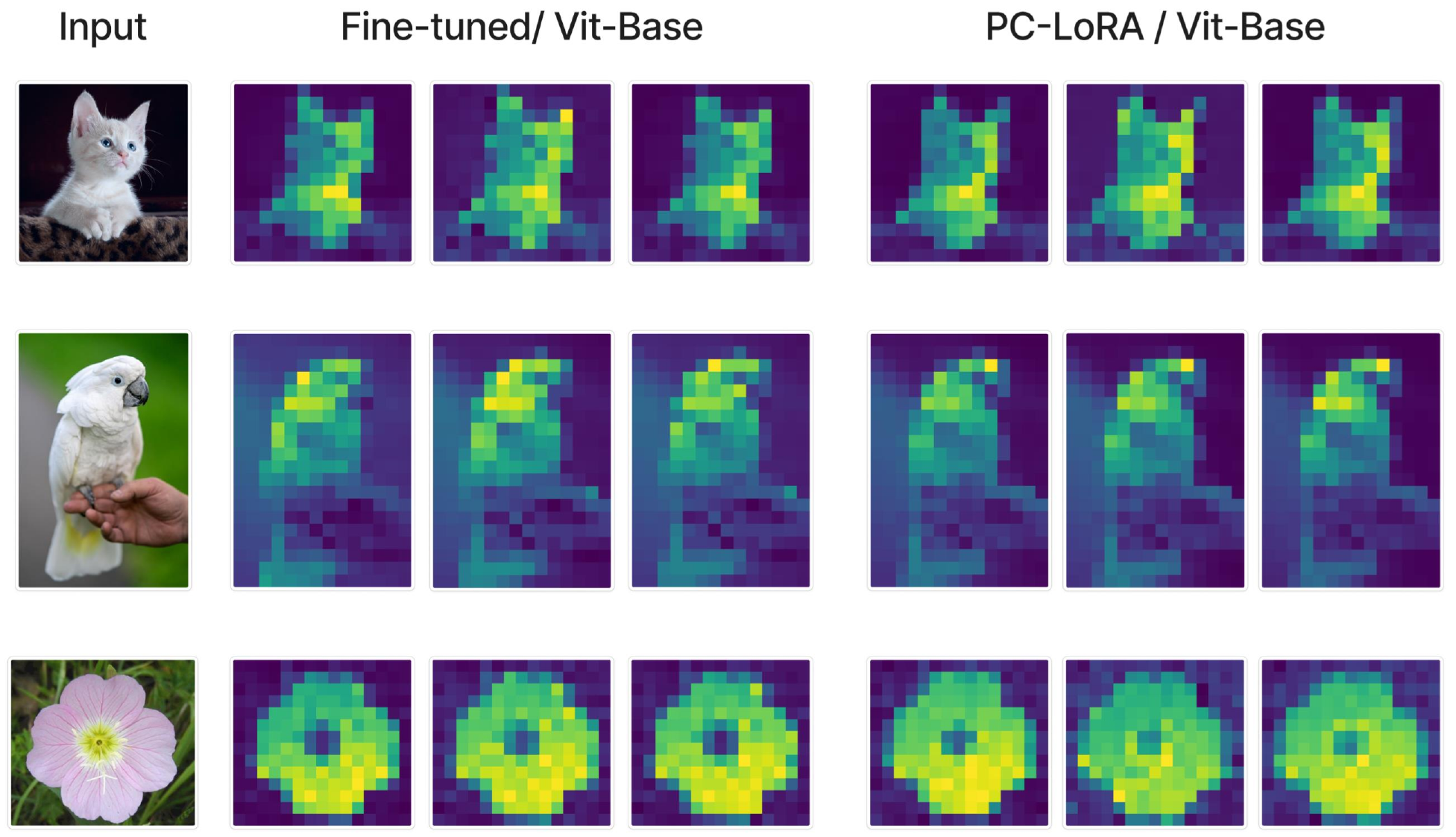}
\vspace{-0.6cm}
\caption{Attention map visualization with [CLS] token: Full-finetuned ViT-B~(85.8M) vs. PC-LoRA ViT-B w/ rank=32~(5.94M). Even with a much smaller model size, our compressed ViT shows comparable attention map quality compared to the full-finetuned ViT-B.
}
\vspace{-0.4cm}
\label{fig:main_attn}
\end{figure}

\noindent
\textbf{Attention Visualization.}
Figure \ref{fig:main_attn} shows a comparison of attention maps across different models and inputs. The left column displays input images of a cat, a parrot, and a flower. The middle column shows the top three attention maps from a ViT Base model fine-tuned on CIFAR-10, highlighting areas of highest activation. The right column features similar attention maps from the model compressed with PC-LoRA, also fine-tuned. Both columns show that the quality of the attention maps is similarly high, indicating that compression does not significantly degrade performance.
However, as detailed in \cref{appendix:c}, it is important to note that the number of heads that effectively contribute to these high-quality attention maps is fewer in the PC-LoRA compressed model. This reduction in effective heads suggests that the compression has been successful at reducing dimensionality without substantially affecting the model's ability to focus on relevant features in the input images.

\section{Conclusion}
In this work, we have explored the ability to combine model compression and parameter-efficient fine-tuning through low-rank adaptation. By progressively attenuating the output of the pre-trained weights and allowing the LoRA weights to bridge the gap of the decayed representation of these weights, our PC-LoRA achieves significant model compression with only a slight drop in performance.

Future work will explore the following potential advancements:
We will improve the Decay Factor Scheduler to enhance compression performance.
Additionally, we plan to refine the initialization of low-rank adapters. Currently, it is initialized with a random Gaussian distribution for \(A\) and zeros for \(B\), essentially starting the compression from a basic setup. 
To enhance the effectiveness of the PC-LoRA method, we will employ a more sophisticated approach by using the results of Singular Value Decomposition (SVD) of pre-trained weights. 
This will serve as the initial information for \(A\) and \(B\), facilitating a progressive compression process.
Moreover, we plan to apply PC-LoRA method to large-scale models and datasets and evaluate our method with other compression strategies, including pruning~\citep{liu2019rethinking} and low-rank factorization~\citep{hsu2022language}.



\noindent
\textbf{Acknowledgement.}
\footnotesize
\noindent
This research was partly supported by Institute of Information \& communications Technology Planning \& Evaluation (IITP)
grant funded by the Korea government (MSIT) (No. RS2022-00187238, Development of Large Korean Language
Model Technology for Efficient Pre-training) and Brian Impact Foundation, a non-profit organization dedicated to the advancement of science and technology for all.  
{
    \small
    \bibliographystyle{cvpr24/ieeenat_fullname}
    \bibliography{main}

\begin{thebibliography}{29}
\providecommand{\natexlab}[1]{#1}
\providecommand{\url}[1]{\texttt{#1}}
\expandafter\ifx\csname urlstyle\endcsname\relax
  \providecommand{\doi}[1]{doi: #1}\else
  \providecommand{\doi}{doi: \begingroup \urlstyle{rm}\Url}\fi

\bibitem[Brown et~al.(2020)Brown, Mann, Ryder, Subbiah, Kaplan, Dhariwal, Neelakantan, Shyam, Sastry, Askell, Agarwal, Herbert-Voss, Krueger, Henighan, Child, Ramesh, Ziegler, Wu, Winter, Hesse, Chen, Sigler, Litwin, Gray, Chess, Clark, Berner, McCandlish, Radford, Sutskever, and Amodei]{brown2020language}
Tom~B. Brown, Benjamin Mann, Nick Ryder, Melanie Subbiah, Jared Kaplan, Prafulla Dhariwal, Arvind Neelakantan, Pranav Shyam, Girish Sastry, Amanda Askell, Sandhini Agarwal, Ariel Herbert-Voss, Gretchen Krueger, Tom Henighan, Rewon Child, Aditya Ramesh, Daniel~M. Ziegler, Jeffrey Wu, Clemens Winter, Christopher Hesse, Mark Chen, Eric Sigler, Mateusz Litwin, Scott Gray, Benjamin Chess, Jack Clark, Christopher Berner, Sam McCandlish, Alec Radford, Ilya Sutskever, and Dario Amodei.
\newblock Language models are few-shot learners, 2020.

\bibitem[Carion et~al.(2020)Carion, Massa, Synnaeve, Usunier, Kirillov, and Zagoruyko]{carion2020endtoend}
Nicolas Carion, Francisco Massa, Gabriel Synnaeve, Nicolas Usunier, Alexander Kirillov, and Sergey Zagoruyko.
\newblock End-to-end object detection with transformers, 2020.

\bibitem[Caron et~al.(2021)Caron, Touvron, Misra, Jégou, Mairal, Bojanowski, and Joulin]{caron2021emerging}
Mathilde Caron, Hugo Touvron, Ishan Misra, Hervé Jégou, Julien Mairal, Piotr Bojanowski, and Armand Joulin.
\newblock Emerging properties in self-supervised vision transformers, 2021.

\bibitem[Clark et~al.(2020)Clark, Luong, Le, and Manning]{clark2020electra}
Kevin Clark, Minh-Thang Luong, Quoc~V. Le, and Christopher~D. Manning.
\newblock Electra: Pre-training text encoders as discriminators rather than generators, 2020.

\bibitem[Dettmers et~al.(2023)Dettmers, Pagnoni, Holtzman, and Zettlemoyer]{dettmers2023qlora}
Tim Dettmers, Artidoro Pagnoni, Ari Holtzman, and Luke Zettlemoyer.
\newblock Qlora: Efficient finetuning of quantized llms, 2023.

\bibitem[Devlin et~al.(2019)Devlin, Chang, Lee, and Toutanova]{devlin2019bert}
Jacob Devlin, Ming-Wei Chang, Kenton Lee, and Kristina Toutanova.
\newblock Bert: Pre-training of deep bidirectional transformers for language understanding, 2019.

\bibitem[Dosovitskiy et~al.(2021)Dosovitskiy, Beyer, Kolesnikov, Weissenborn, Zhai, Unterthiner, Dehghani, Minderer, Heigold, Gelly, Uszkoreit, and Houlsby]{dosovitskiy2021image}
Alexey Dosovitskiy, Lucas Beyer, Alexander Kolesnikov, Dirk Weissenborn, Xiaohua Zhai, Thomas Unterthiner, Mostafa Dehghani, Matthias Minderer, Georg Heigold, Sylvain Gelly, Jakob Uszkoreit, and Neil Houlsby.
\newblock An image is worth 16x16 words: Transformers for image recognition at scale, 2021.

\bibitem[Heo et~al.(2018)Heo, Lee, Yun, and Choi]{heo2018knowledge}
Byeongho Heo, Minsik Lee, Sangdoo Yun, and Jin~Young Choi.
\newblock Knowledge transfer via distillation of activation boundaries formed by hidden neurons, 2018.

\bibitem[Hinton et~al.(2015)Hinton, Vinyals, and Dean]{hinton2015distilling}
Geoffrey Hinton, Oriol Vinyals, and Jeff Dean.
\newblock Distilling the knowledge in a neural network, 2015.

\bibitem[Houlsby et~al.(2019)Houlsby, Giurgiu, Jastrzebski, Morrone, de~Laroussilhe, Gesmundo, Attariyan, and Gelly]{houlsby2019parameter}
Neil Houlsby, Andrei Giurgiu, Stanislaw Jastrzebski, Bruna Morrone, Quentin de Laroussilhe, Andrea Gesmundo, Mona Attariyan, and Sylvain Gelly.
\newblock Parameter-efficient transfer learning for nlp, 2019.

\bibitem[Hsu et~al.(2022)Hsu, Hua, Chang, Lou, Shen, and Jin]{hsu2022language}
Yen-Chang Hsu, Ting Hua, Sungen Chang, Qian Lou, Yilin Shen, and Hongxia Jin.
\newblock Language model compression with weighted low-rank factorization, 2022.

\bibitem[Hu et~al.(2021)Hu, Shen, Wallis, Allen-Zhu, Li, Wang, Wang, and Chen]{hu2021lora}
Edward~J. Hu, Yelong Shen, Phillip Wallis, Zeyuan Allen-Zhu, Yuanzhi Li, Shean Wang, Lu Wang, and Weizhu Chen.
\newblock Lora: Low-rank adaptation of large language models, 2021.

\bibitem[Jacob et~al.(2017)Jacob, Kligys, Chen, Zhu, Tang, Howard, Adam, and Kalenichenko]{jacob2017quantization}
Benoit Jacob, Skirmantas Kligys, Bo Chen, Menglong Zhu, Matthew Tang, Andrew Howard, Hartwig Adam, and Dmitry Kalenichenko.
\newblock Quantization and training of neural networks for efficient integer-arithmetic-only inference, 2017.

\bibitem[Krizhevsky(2009)]{Krizhevsky2009LearningML}
Alex Krizhevsky.
\newblock Learning multiple layers of features from tiny images, 2009.

\bibitem[Lester et~al.(2021)Lester, Al-Rfou, and Constant]{Lester2021ThePO}
Brian Lester, Rami Al-Rfou, and Noah Constant.
\newblock The power of scale for parameter-efficient prompt tuning.
\newblock In \emph{Conference on Empirical Methods in Natural Language Processing}, 2021.

\bibitem[Li and Liang(2021)]{li2021prefixtuning}
Xiang~Lisa Li and Percy Liang.
\newblock Prefix-tuning: Optimizing continuous prompts for generation, 2021.

\bibitem[Liu et~al.(2021)Liu, Zheng, Du, Ding, Qian, Yang, and Tang]{Liu2021GPTUT}
Xiao Liu, Yanan Zheng, Zhengxiao Du, Ming Ding, Yujie Qian, Zhilin Yang, and Jie Tang.
\newblock Gpt understands, too.
\newblock \emph{ArXiv}, abs/2103.10385, 2021.

\bibitem[Liu et~al.(2019{\natexlab{a}})Liu, Ott, Goyal, Du, Joshi, Chen, Levy, Lewis, Zettlemoyer, and Stoyanov]{liu2019roberta}
Yinhan Liu, Myle Ott, Naman Goyal, Jingfei Du, Mandar Joshi, Danqi Chen, Omer Levy, Mike Lewis, Luke Zettlemoyer, and Veselin Stoyanov.
\newblock Roberta: A robustly optimized bert pretraining approach, 2019{\natexlab{a}}.

\bibitem[Liu et~al.(2019{\natexlab{b}})Liu, Sun, Zhou, Huang, and Darrell]{liu2019rethinking}
Zhuang Liu, Mingjie Sun, Tinghui Zhou, Gao Huang, and Trevor Darrell.
\newblock Rethinking the value of network pruning, 2019{\natexlab{b}}.

\bibitem[Loshchilov and Hutter(2017)]{loshchilov2017sgdr}
Ilya Loshchilov and Frank Hutter.
\newblock Sgdr: Stochastic gradient descent with warm restarts, 2017.

\bibitem[Loshchilov and Hutter(2019)]{loshchilov2019decoupled}
Ilya Loshchilov and Frank Hutter.
\newblock Decoupled weight decay regularization, 2019.

\bibitem[Maas et~al.(2011)Maas, Daly, Pham, Huang, Ng, and Potts]{maas-etal-2011-learning}
Andrew~L. Maas, Raymond~E. Daly, Peter~T. Pham, Dan Huang, Andrew~Y. Ng, and Christopher Potts.
\newblock Learning word vectors for sentiment analysis.
\newblock In \emph{Proceedings of the 49th Annual Meeting of the Association for Computational Linguistics: Human Language Technologies}, pages 142--150, Portland, Oregon, USA, 2011. Association for Computational Linguistics.

\bibitem[Park et~al.(2019)Park, Kim, Lu, and Cho]{park2019relational}
Wonpyo Park, Dongju Kim, Yan Lu, and Minsu Cho.
\newblock Relational knowledge distillation, 2019.

\bibitem[Passalis and Tefas(2019)]{passalis2019learning}
Nikolaos Passalis and Anastasios Tefas.
\newblock Learning deep representations with probabilistic knowledge transfer, 2019.

\bibitem[Radford et~al.(2021)Radford, Kim, Hallacy, Ramesh, Goh, Agarwal, Sastry, Askell, Mishkin, Clark, Krueger, and Sutskever]{radford2021learning}
Alec Radford, Jong~Wook Kim, Chris Hallacy, Aditya Ramesh, Gabriel Goh, Sandhini Agarwal, Girish Sastry, Amanda Askell, Pamela Mishkin, Jack Clark, Gretchen Krueger, and Ilya Sutskever.
\newblock Learning transferable visual models from natural language supervision, 2021.

\bibitem[Romero et~al.(2015)Romero, Ballas, Kahou, Chassang, Gatta, and Bengio]{romero2015fitnets}
Adriana Romero, Nicolas Ballas, Samira~Ebrahimi Kahou, Antoine Chassang, Carlo Gatta, and Yoshua Bengio.
\newblock Fitnets: Hints for thin deep nets, 2015.

\bibitem[Vaswani et~al.(2023)Vaswani, Shazeer, Parmar, Uszkoreit, Jones, Gomez, Kaiser, and Polosukhin]{vaswani2023attention}
Ashish Vaswani, Noam Shazeer, Niki Parmar, Jakob Uszkoreit, Llion Jones, Aidan~N. Gomez, Lukasz Kaiser, and Illia Polosukhin.
\newblock Attention is all you need, 2023.

\bibitem[Wang et~al.(2021)Wang, Xie, Li, Fan, Song, Liang, Lu, Luo, and Shao]{wang2021pyramid}
Wenhai Wang, Enze Xie, Xiang Li, Deng-Ping Fan, Kaitao Song, Ding Liang, Tong Lu, Ping Luo, and Ling Shao.
\newblock Pyramid vision transformer: A versatile backbone for dense prediction without convolutions, 2021.

\bibitem[Zagoruyko and Komodakis(2017)]{zagoruyko2017paying}
Sergey Zagoruyko and Nikos Komodakis.
\newblock Paying more attention to attention: Improving the performance of convolutional neural networks via attention transfer, 2017.

\end{thebibliography}
}

\clearpage
\appendix

\paragraph{\LARGE{Appendix}}

\section{Related Works}

\subsection{Low-Rank Adaptation (LoRA)} LoRA fine-tuning technique involves adding a few trainable parameters while keeping the original model parameters fixed. This is done by modifying the pre-trained weight matrix \(W_0\) with a small, low-rank update \(\Delta W = BA\), where \(B\) and \(A\) are matrices of smaller dimensions. During training, only \(A\) and \(B\) are updated, with \(W_0\) remaining unchanged. The method calculates the new output \(h\) by adding \(W_0x\) and \(BAx\) together. Our approach also includes adding a bias term and scheduling the reduction of certain parameters to improve model efficiency and restoration.

\subsection{Feature-based Knowledge Distillation}
Knowledge distillation (KD) encompasses three principal methodologies: Response-based~\cite{hinton2015distilling}, Feature-based~\cite{romero2015fitnets}, and Relation-based KD~\cite{park2019relational}. Each specializes in different aspects of knowledge transfer from a teacher to a student model. Given that PC-LoRA method is layer-wise, we have utilized Feature-based KD to leverage the intermediate layers' semantic information in the training process.
Feature-based KD focuses on the use of feature maps from both the student and teacher models, enhancing the learning process with a strategic focus on the intermediate representations. A distance function, such as Mean Squared Error (MSE) loss~\cite{romero2015fitnets, zagoruyko2017paying, heo2018knowledge} or Kullback-Leibler divergence loss~\cite{passalis2019learning} is used to quantify the similarity of the matched features. 
Our adapter modules generate feature maps that match the size of the original model's, eliminating the need for alterations required by other feature-based knowledge distillation methods and thereby avoiding associated losses.

\label{appendix:related}
\section{Training Details}\label{appendix:a}
\subsection{Datasets}
The IMDb dataset includes 50,000 movie reviews, evenly divided between training and testing sets. The CIFAR-10 dataset contains 60,000 color images in 10 classes, split into 50,000 for training and 10,000 for testing.
\subsection{Settings for Training}
For the optimization of the PC-LoRA method, we used AdamW ~\citep{loshchilov2019decoupled} optimizer, exploring learning rates within {1e-2, 1e-5}. In terms of learning rate scheduling, we adopted CosineAnnealingLR ~\citep{loshchilov2017sgdr}, which is set with a minimum value of 0. We used the batch size to be 64 for CIFAR-10 image classification task and 20 for IMDb text classification task. The image classification was trained for 62,500 iterations, and the text classification task, 100,000 iterations.
Additionally, all experiments were conducted on RTX 5000 GPUs using PyTorch version 2.1, Python 3.10, and CUDA 11.8.0.

\section{More experimental analysis}\label{appendix:D}

\subsection{Loss Scale factor, $\alpha$}\label{appendix:loss_scale_factor}

We differed Decay Factor Scheduler with sine: \cref{tab:loss_ratios_performance1}, linear: \cref{tab:loss_ratios_performance2}, and 1-cosine: \cref{tab:loss_ratios_performance3}. For each table, we conducted an ablation study for the loss scale factor $\alpha$ value in Equation \ref{eq:4}. It's important to note that an alpha value of 1 implies training solely with the \(L_{\text{task}}\) task loss and setting the $\alpha$ value to 0 results in no fine-tuning, which explains our decision to compare a $\alpha$ values ranging from 0.2 to 1.0.  
Our results indicate that, generally, performances are better with $\alpha$ values lower than 1. Specifically, in the CIFAR-10 task, there was an average performance increase of 1.25\%p, while in the IMDb task, there was an increase of 2.64\%p. This tendency was observed regardless of the type of Decay Factor Scheduler, demonstrating that lower $\alpha$ values enhance performance more effectively than an $\alpha$ of 1.

\begin{table}[h]
\centering
\begin{tabular}{lccccc}
\toprule
& \multicolumn{5}{c}{\small {\textbf{$\alpha$ Acc(\%)}}} \\ 
\cline{2-6} 
\textbf{Model}   & \textbf{0.2} & \textbf{0.4} & \textbf{0.6} & \textbf{0.8}& \textbf{1.0}\\ \toprule
ViT-B            & \textbf{95.04}               & 94.84                & 93.76                & 93.64  & 93.88                        \\
ViT-B/Clip       & \textbf{94.12}               & 93.84                & 93.36                & 93.60   & 93.36                       \\ 
ViT-B/DINO       & \textbf{95.20}               & 94.52                & 93.60                & 93.88    & 93.72                      \\ 
\midrule
BERT             & \textbf{92.67}               & 92.24                & 91.33                & 90.78   & 89.71                       \\ 
ELECTRA          & \textbf{92.86}               & 91.82                & 91.34                & 90.83   & 90.10                       \\ 
RoBERTa          & \textbf{92.51}              & 92.11                & 91.74                & 90.78   & 90.32                       \\ \toprule
\end{tabular}
\caption{Decay Scheduler Sine}
\label{tab:loss_ratios_performance1}
\end{table}

\begin{table}[h]
\centering
\begin{tabular}{lccccc}
\toprule
& \multicolumn{5}{c}{\small {\textbf{$\alpha$ Acc(\%)}}} \\ 
\cline{2-6} 
\textbf{Model}   & \textbf{0.2} & \textbf{0.4} & \textbf{0.6} & \textbf{0.8} & \textbf{1.0}\\ \toprule
ViT-B            & \textbf{95.00}               & 94.40                & 93.72                & 93.16    & 92.60                      \\
ViT-B/Clip       & 93.44               & 92.96                & 92.60                & \textbf{93.60}   & \textbf{93.60}                       \\ 
ViT-B/DINO       & \textbf{94.2}               & 93.96                & 93.64                & 93.80   & 93.00                       \\ 
\midrule
BERT             & \textbf{92.78}               & 92.08                & 91.41                & 90.61     & 89.92                     \\ 
ELECTRA          & \textbf{92.54}               & 91.94                & 91.22                & 90.62     & 90.13                     \\ 
RoBERTa          & \textbf{92.59}               & 92.30                & 91.44                & 91.26     &90.03                     \\ \toprule
\end{tabular}
\caption{Decay Scheduler linear}
\label{tab:loss_ratios_performance2}
\end{table}

\begin{table}[h]
\centering
\begin{tabular}{lccccc}
\toprule
& \multicolumn{5}{c}{\small {\textbf{$\alpha$ Acc(\%)}}} \\ 
\cline{2-6} 
\textbf{Model}   & \textbf{0.2} & \textbf{0.4} & \textbf{0.6} & \textbf{0.8} & \textbf{1.0}\\ \toprule
ViT-B            & \textbf{94.16}                & 93.56                & 93.28                & 93.04    & 92.36                      \\
ViT-B/Clip       & 93.40                & \textbf{93.72}                & 93.20                & 92.92      & 92.44                    \\ 
ViT-B/DINO       & \textbf{94.16}                & 93.52                & 93.00                & 93.56       & 92.84                   \\ 
\midrule
BERT             & \textbf{92.67}                & 91.55                & 90.98                & 90.24    & 89.58                      \\ 
ELECTRA          & \textbf{92.42}               & 92.05                & 91.31                & 90.34    & 89.95                      \\ 
RoBERTa          & 91.84               & \textbf{92.00}                & 91.17                & 90.91    & 89.38                      \\ \toprule
\end{tabular}
\caption{Decay Scheduler 1-Cosine}
\label{tab:loss_ratios_performance3}
\end{table}

\FloatBarrier

\subsection{Feature Knowledge Distillation}
\FloatBarrier
\begin{table}[h]
\centering
\begin{tabular}{lcc}
\toprule
& \multicolumn{2}{c}{\small {\textbf{Acc(\%)}}} \\ 
\cline{2-3} 
\textbf{Model}   & {\small \textbf{w/o $\loss{featKD}$}} & {\small \textbf{w/ $\loss{featKD}$}} \\ \toprule
ViT-B            & 93.88                & \textbf{95.04}                          \\
ViT-B/Clip       & 93.36                & \textbf{94.12}                          \\
ViT-B/DINO       & 93.72                & \textbf{95.20}                           \\
\midrule
BERT             & 89.92                & \textbf{92.78}                            \\
ELECTRA          & 90.13                & \textbf{92.86}                            \\
RoBERTa          & 90.32                & \textbf{92.59}                            \\ \toprule
\end{tabular}
\caption{
The performance of various pre-trained models with and without the 
feature knowledge distillation(FKD) was evaluated using the CIFAR-10 and IMDB benchmarks.
w/o FKD means using only the task loss, $\loss{task}$.
}
\label{tab:fd_loss_performance}
\end{table}

\FloatBarrier
\subsection{Decay Scheduler Styles}
\begin{table}[h]
\small
\centering
\begin{tabular}{lccc}
\toprule
& \multicolumn{3}{c}{\small {\textbf{with FKD Acc(\%)}}} \\ %
\cline{2-4} 
\textbf{Model}   & \textbf{Linear} & \textbf{1-Cosine} & \textbf{Sine}\\ \toprule
ViT-B            & 95.00                & 94.16                & \textbf{95.04}                          \\
ViT-B/Clip           & 93.96                & 93.72                & \textbf{94.12}                          \\
ViT-B/DINO           & 94.20                & 94.16                & \textbf{95.20}                          \\
\midrule
BERT           & \textbf{92.78}                & 92.67                & 92.67                          \\
ELECTRA           & 92.54                & 92.42                & \textbf{92.86}                          \\
RoBERTa           & \textbf{92.59}                & 92.00                & 92.51                          \\ \toprule
\end{tabular}
\caption{
The performance comparison of various decay schedulers on pre-trained models with FKD.}
\label{tab:decay_scheduler_performance}
\end{table}

\FloatBarrier

\section{Compression comparison}\label{appendix:E}
We've conducted a comparison of model compression using PC-LoRA on both ViT\_Base and BERT\_Base architectures. Compression was performed across a broad spectrum of rank values, including 16, 32, 64, 128, and 256, to achieve models of various sizes. Upon comparing the performance of the compressed models with their corresponding baseline models of similar size, both ViT and BERT architectures demonstrated an overall improvement in performance. 
\subsection{ViT compression with PC-LoRA}
\begin{table}[h]
\small
\centering
\renewcommand{\arraystretch}{1.2} 
\begin{tabular}{lcc}
\toprule
\textbf{Model}         & \textbf{Parameters (M)} & \textbf{ Accuracy (\%)} \\ \toprule
ViT\_Tiny              & 5.53                   & 94.66                  \\ 
ViT\_Small             & 21.67                   & 97.01                     \\
ViT\_Base              & 85.81                  & 98.80                   \\ \midrule
ViT\_Base r=32         & 5.59                    & 95.04                     \\ 
ViT\_Base r=64         & 9.87                   & 96.44                  \\ 
ViT\_Base r=128        & 19.75                  & 97.08                  \\ 
ViT\_Base r=256        & 38.62                   & 97.56                  \\ \bottomrule
\end{tabular}
\caption{The performance of models based on different size of ViT vs PC-LoRA diversly compressed ViT\-Base : Fine-tuned with CIFAR-10. The table represents specific model sizes and performance as shown in Figure \ref{fig:vit_line_plot}.  ViT\_Tiny, Small, and Base represent the baseline models, while ViT\_Base r=32, 64, 128, 256 indicate models that have been compressed to various sizes through PC-LoRA, applied to the ViT\_Base model.}
\label{tab:vit plot table}
\end{table}

\FloatBarrier

\subsection{BERT compression with PC-LoRA}
\begin{table}[h]
\small
\centering
\renewcommand{\arraystretch}{1.2} 
\begin{tabular}{lcc}
\toprule
\textbf{Model}         & \textbf{Parameters (M)} & \textbf{ Accuracy (\%)} \\ \toprule
BERT\_Small              & 28.77                   & 92.27                  \\ 
BERT\_Medium             & 41.37                   & 92.38                     \\ 
BERT\_Base              & 109.48                  & 94.12                   \\ \midrule
BERT\_Base r=16         & 26.65                    & 92.40                     \\ 
BERT\_Base r=32         & 29.34                   & 92.67                  \\ 
BERT\_Base r=128        & 45.49                  & 93.12                  \\ 
BERT\_Base r=256        & 67.02                   & 93.36                  \\ \bottomrule
\end{tabular}
\caption{The performance of models based on different sizes of BERT vs PC-LoRA diversly compressed BERT\_Base: Fine-tuned with IMDb. The table represents specific model sizes and performance as shown in Figure \ref{fig:bert_graph}.  BERT\_Small, Medium, and Base represent the baseline models, while BERT\_Base r=16, 32, 128, 256 indicate models that have been compressed to various sizes through PC-LoRA, applied to the BERT\_Base model.}
\label{tab:bert plot table}
\end{table}
\begin{figure}[h]
\centering
\includegraphics[width=0.98\linewidth,height=0.66\linewidth]{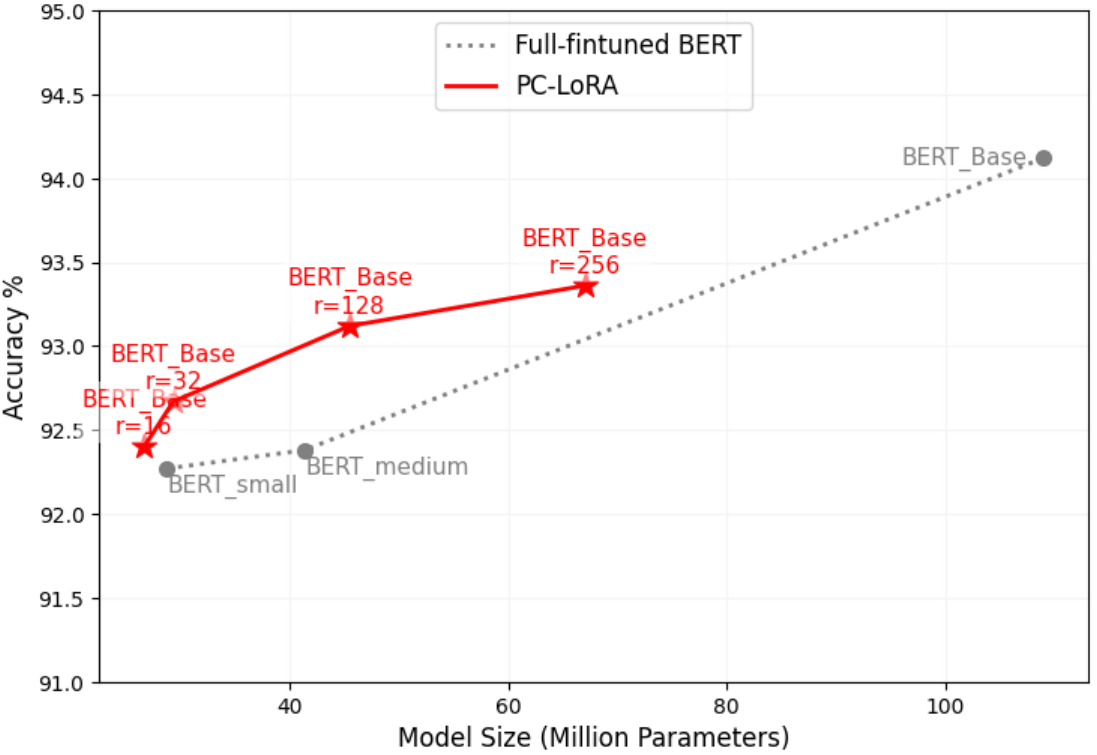}
\caption{The performance comparisons based on different compression ratios of PC-LoRA using BERT~\cite{devlin2019bert} compared to the fully finetuned BERT-B on IMDb.}
\label{fig:bert_graph}
\end{figure}

\FloatBarrier

\section{Models Information}\label{appendix:b}
We employed base versions of various pre-trained models. 
We used Bert-base~\citep{devlin2019bert}, RoBERTa-base~\citep{liu2019roberta}, and ELECTRA-base~\citep{clark2020electra} for text classification task, which were sourced from the HuggingFace Transformers library.
For image classification task , we used ViT-base~\citep{dosovitskiy2021image}, ViT-base/DINO~\citep{caron2021emerging}, Clip-ViT-base~\citep{radford2021learning}, which incorporates the ResNet-50 architecture, sourced from the 'timm' library.

\subsection*{Download Links for pre-trained Models}
\begin{enumerate}
    \item ViT-base/DINO : \href{https://huggingface.co/timm/vit_base_patch16_224.dino}{timm/vit\_base\_patch16\_224.dino}
    \item ViT-base/Clip : \href{https://huggingface.co/timm/vit_base_patch16_clip_224.openai}{timm/vit\_base\_patch16\_clip\_224.openai}
    \item Bert-base : \href{https://huggingface.co/bert-base-uncased}{bert-base-uncased}
    \item RoBERTa-base IMDb : \href{https://huggingface.co/FacebookAI/roberta-base}{FacebookAI/roberta-base}
    \item ELECTRA-base: \href{https://huggingface.co/google/electra-base-discriminator}{google/electra-base-discriminator}
\end{enumerate}

\subsection*{Download Links for Fine-tuned Models}

\begin{enumerate}
    \item Bert-base IMDb : \href{https://huggingface.co/nikitakapitan/bert-base-uncased-finetuned-imdb}{nikitakapitan/bert-base-uncased-finetuned-imdb}
    \item RoBERTa-base IMDb : \href{https://huggingface.co/aychang/roberta-base-imdb}{aychang/roberta-base-imdb}
    \item ELECTRA-base: \href{https://huggingface.co/pig4431/IMDB_ELECTRA_5E}{pig4431/IMDb\_ELECTRA\_5E}
\end{enumerate}

For these two models, ViT-base/DINO and ViT-base/CLIP, we have conducted fine-tuning on CIFAR-10 by searching for the optimal hyperparameters
\\
\\
\\
\\
\\
\\

\section{Attention Maps}\label{appendix:c}

\begin{figure}[h]
\centering
\includegraphics[width=\linewidth]{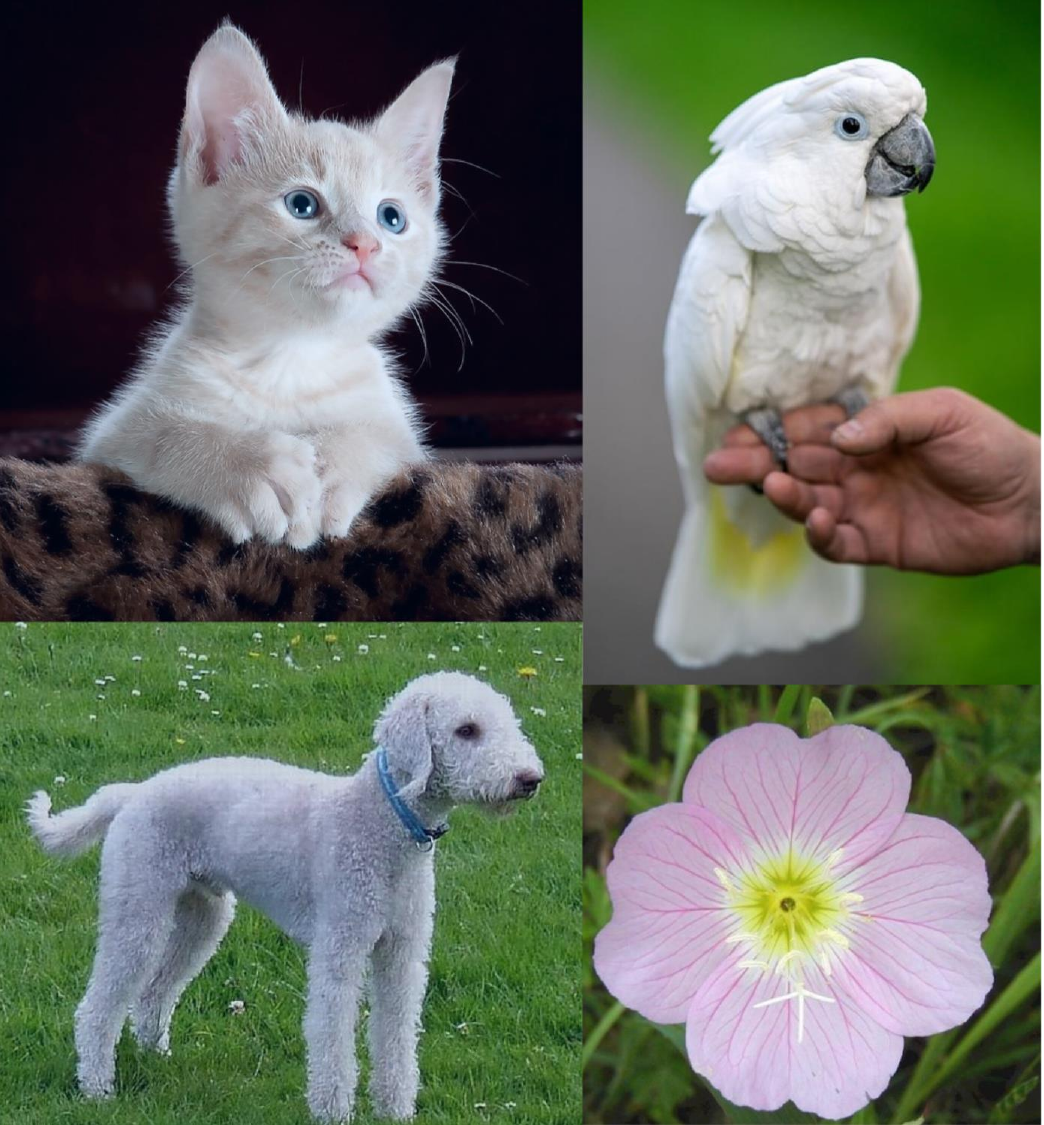}
\caption{Top Left: a cat. Top Right: a white parrot. Bottom Left: a dog. Bottom Right: a flower}
\label{fig:all}
\end{figure}

\begin{figure}[h]
\centering
\includegraphics[width=\linewidth]{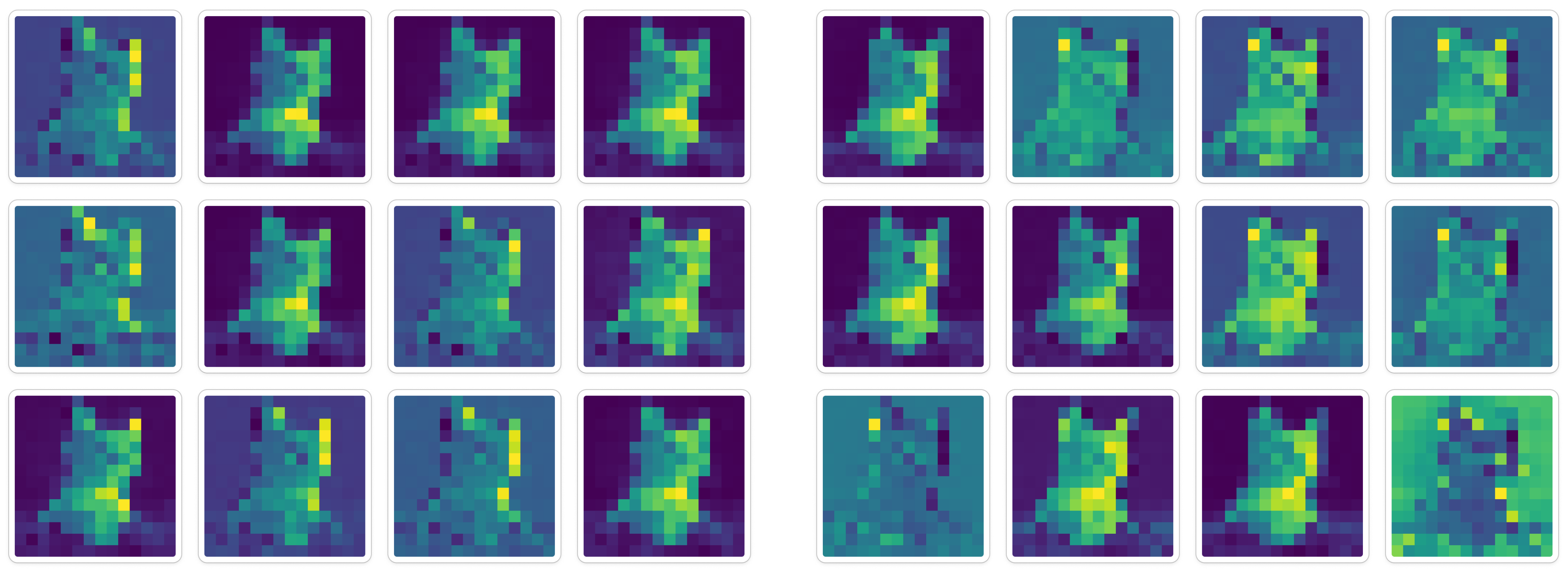}
\caption{Left cluster of 12 images: attention maps of ViT Base fine-tuned on CIFAR-10 for a cat. Right cluster of 12 images: attention maps from the same model compressed with PC-LoRA technique to Rank 32.}
\label{fig:cat_attn}
\end{figure}

\begin{figure}[h]
\centering
\includegraphics[width=\linewidth]{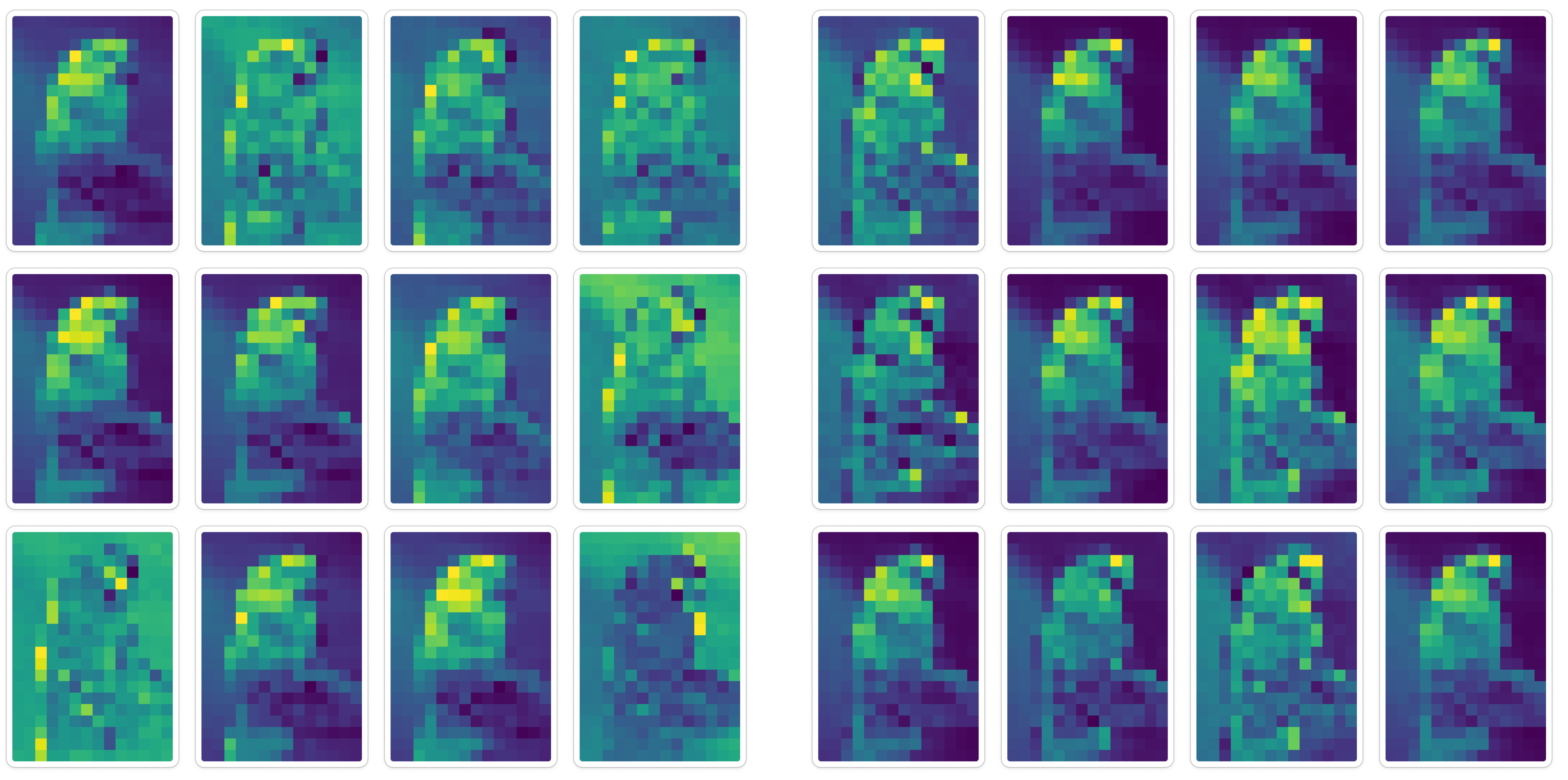}
\caption{Left cluster of 12 images: attention maps of a ViT Base fine-tuned on CIFAR-10 for a white parrot. Right cluster of 12 images: attention maps from the same model compressed with PC-LoRA technique to Rank 32.}
\label{fig:bird_attn}
\end{figure}

\begin{figure}[h]
\centering
\includegraphics[width=\linewidth]{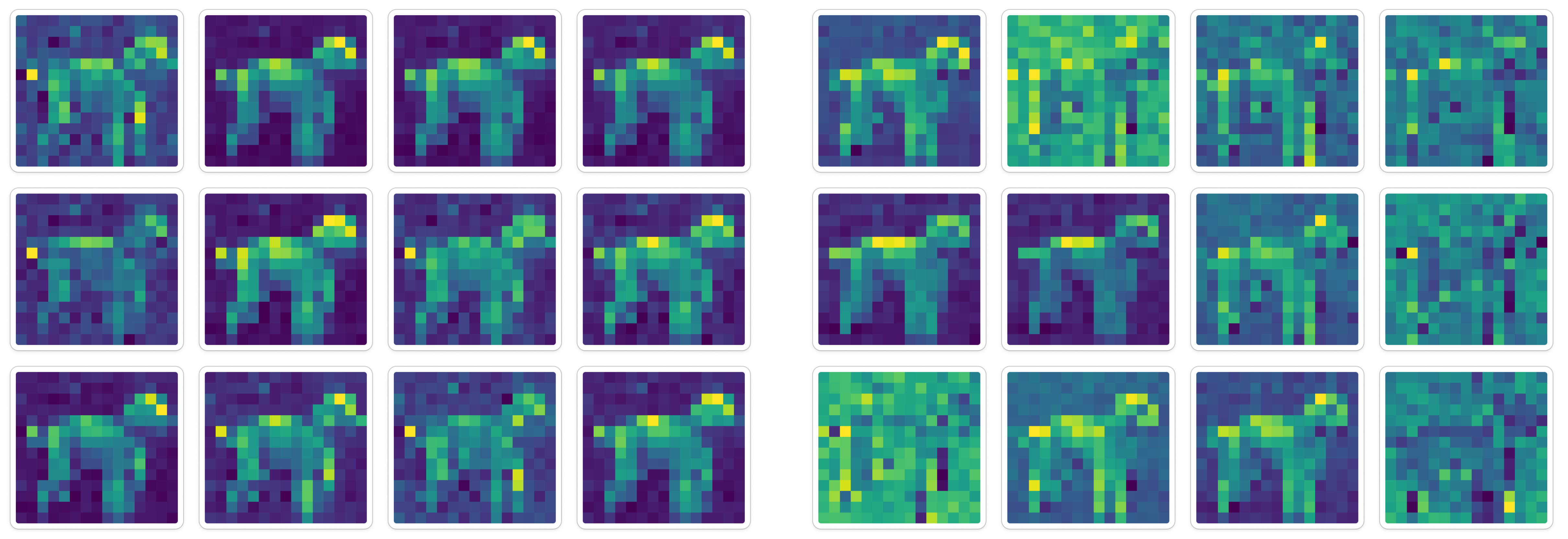}
\caption{Left cluster of 12 images: attention maps of a ViT Base fine-tuned on CIFAR-10 for a dog. Right cluster of 12 images: attention maps from the same model compressed with PC-LoRA technique to Rank 32.}
\label{fig:dog_attn}
\end{figure}

\begin{figure}[h]
\centering
\includegraphics[width=\linewidth]{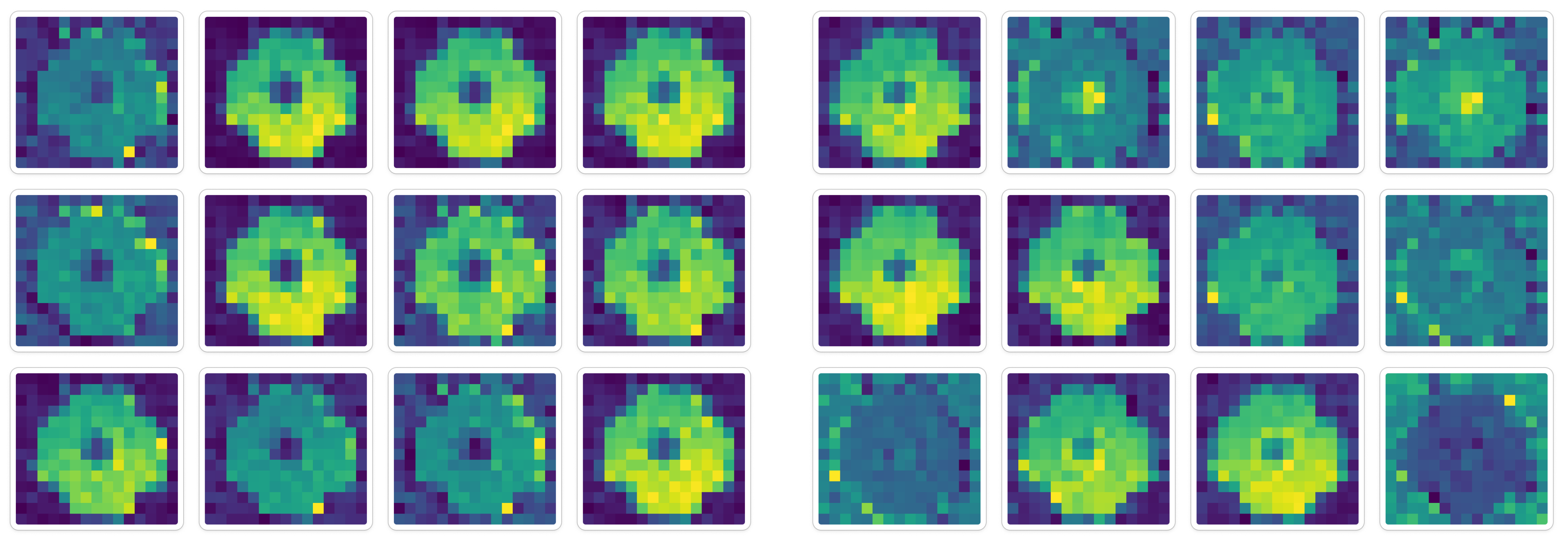}
\caption{Left cluster of 12 images: attention maps of a ViT Base fine-tuned on CIFAR-10 for a flower. Right cluster of 12 images: attention maps from the same model compressed with PC-LoRA technique to Rank 32.}
\label{fig:flower_attn}
\end{figure}

\FloatBarrier


\end{document}